\documentclass{IEEE_lsens}
\usepackage{textcomp}

\usepackage{multirow}

\usepackage[noadjust]{cite}

\ifCLASSINFOpdf
\else
\fi
\usepackage[T1]{fontenc} 
\usepackage{amsmath}
\usepackage{graphicx}

\usepackage[cmintegrals]{newtxmath}
\usepackage{bm}

\usepackage{array}
\usepackage{url}
\ifCLASSINFOpdf
\else
\fi
\providecommand{\hypersetup}[1]{\relax}

\hypersetup{pdftitle={Reliability-Aware Reconstruction of V-Shaped Surfaces with a Low-Cost Multi-Zone SPAD Sensor},
pdfsubject={Typesetting},
pdfauthor={Yohanssen Pratama},
pdfkeywords={Class, IEEE, IEEE\_lsens, IEEE Sensors Letters, LaTeX, Typesetting, TeX}}
\begin{document}


\IEEELSENSarticlesubject{Sensor Applications}

%
\title{Reliability-Aware Reconstruction of V-Shaped Surfaces with a Low-Cost Multi-Zone SPAD Sensor}

%
\author{\IEEEauthorblockN{Yohanssen~Pratama\IEEEauthorrefmark{1}}
\IEEEauthorblockA{\IEEEauthorrefmark{1}
Institute for International Strategy and Emerging Technologies,
Tokyo International University, 1-13-1 Matoba-Kita, Kawagoe, 350-1197, Japan}
\thanks{Corresponding author: Y. Pratama (e-mail: ypratama@tiu.ac.jp).}}
%
%
%


\IEEEtitleabstractindextext{%
\begin{abstract}
Low-cost multi-zone single-photon avalanche diode (SPAD) sensors provide
compact time-of-flight measurements but are limited by sparse sampling and
unreliable depth estimates near geometric discontinuities. This letter
presents a reliability-aware reconstruction framework for V-shaped surfaces
using two complementary TMF8821 acquisition configurations. Point-wise
reliability is learned from sensor confidence, transient peak amplitude, and
peak sharpness, and is incorporated into a graph-based depth optimization.
The reconstructed geometry is evaluated by robust two-plane fitting and a
perpendicular cross-section. Experiments on $60^\circ$, $90^\circ$, and
$120^\circ$ targets reduce the mean V-angle error from $11.21^\circ$ to
$9.66^\circ$ ($13.8\%$). For the representative $90^\circ$ target, the
combined point-to-plane RMSE decreases from $14.56$~mm to $13.65$~mm.
\end{abstract}

}


\maketitle

\section{Introduction}

Time-of-flight (ToF) sensing is widely used for three-dimensional (3-D)
perception. Single-photon avalanche diode (SPAD) sensors are attractive
because of their single-photon sensitivity and high temporal resolution
\cite{bruschini2019spad,scholes2023limits}. Computational single-photon
imaging has further demonstrated recovery of depth and reflectivity from
sparse photon measurements by exploiting spatial and transient information
\cite{kirmani2014first,shin2016photon,altmann2016robust}.

Accurate reconstruction remains difficult near sharp depth transitions,
where mixed returns can produce erroneous or ``flying'' measurements
\cite{chugunov2021masktof}. This is especially relevant for V-shaped
surfaces, whose two planar faces meet at a sharp fold. Compact multi-zone
dToF sensors such as the TMF8821 integrate SPAD detection, TDC circuitry,
histogram processing, and distance/confidence output in a low-cost module
\cite{amsosram_tmf8821}; however, their sparse spatial sampling limits direct
surface recovery.

This letter proposes a reliability-aware reconstruction framework that
aggregates two complementary acquisition configurations and learns
point-wise reliability from confidence, transient peak amplitude, and peak
sharpness. Reliability-weighted graph optimization refines the measured
depth while preserving transient and local geometric consistency. The main
contribution is the use of transient-derived, learnable reliability to guide
geometric refinement of sparse low-cost SPAD measurements. Experiments on
$60^\circ$, $90^\circ$, and $120^\circ$ V-shaped targets reduce the mean
angular error from $11.21^\circ$ to $9.66^\circ$ ($13.8\%$).

\section{Reliability-Aware SPAD Surface Reconstruction}

\begin{figure}[t]
    \centering
    \includegraphics[width=\columnwidth]{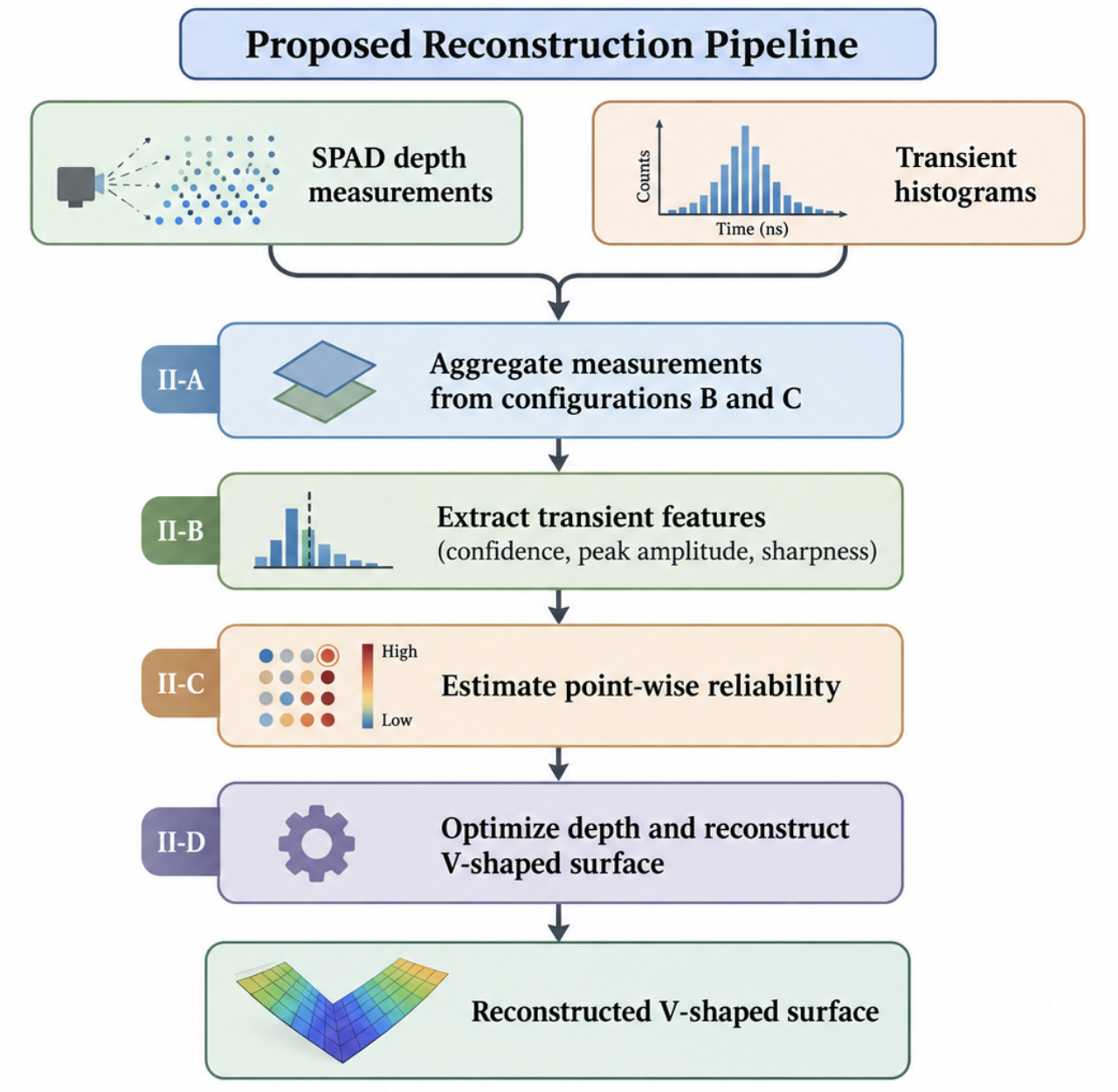}
    \caption{Overview of the proposed reliability-aware V-shaped surface reconstruction pipeline.}
    \label{fig:pipeline}
\end{figure}

The proposed framework reconstructs V-shaped surfaces from sparse
multi-zone SPAD measurements by jointly exploiting depth and transient
information. Measurements acquired under two sensor configurations are
first aggregated into a common point set. Transient-derived features are
then used to estimate the reliability of each depth measurement.
Finally, the measured depth is refined through a graph-based optimization
that balances measurement fidelity, local geometric consistency, and
transient peak-position consistency. The overall processing pipeline is shown in Fig.~\ref{fig:pipeline}.

\subsection{Multi-Configuration SPAD Acquisition}

The measurement setup employs a TMF8821 multi-zone direct
time-of-flight (dToF) sensor to acquire both depth and transient photon
histograms from the target surface. The sensor is operated using two
custom acquisition configurations, denoted as Configurations B and C,
to provide complementary spatial sampling of the scene. Both
configurations are defined over a $16\times10$ logical sampling grid and
are executed sequentially using 20 measurement masks. Each acquisition
frame uses one measurement per mask after a 5-ms settling interval.
Measurements with confidence below 75 or nonpositive range are discarded
before reconstruction.

\begin{figure}[t]
    \centering
    \includegraphics[width=\columnwidth]{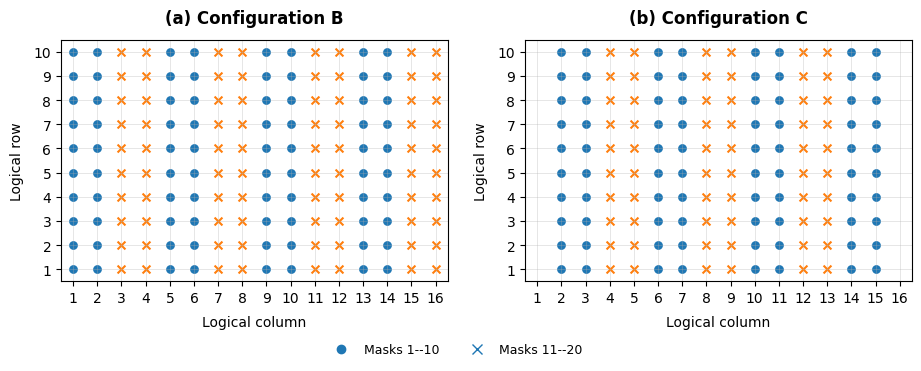}
    \caption{Spatial sampling patterns of the custom TMF8821 configurations:
(a) Configuration B and (b) horizontally shifted Configuration C.}
    \label{fig:sampling}
\end{figure}

Configuration B covers the complete $16\times10$ grid, whereas
Configuration C applies a one-column horizontal shift to provide
complementary measurements under a different sampling pattern. The two
configurations therefore observe the same target geometry with different
spatial support. Valid measurements from both configurations are aggregated into a common
point set before reliability estimation and depth optimization. The two
sampling patterns are shown in Fig.~\ref{fig:sampling}.

Each valid sample provides range, confidence, and a temporal histogram.
Before optimization, samples below the 20th confidence percentile or
15th photon-count percentile, above the 90th FWHM percentile, farther
than 12~mm from the local median depth, or within the most isolated
10\% of samples are removed.

After FoV calibration, each reported distance $z_i$ is embedded in the
Cartesian parameterization used for reconstruction according to

\begin{equation}
\begin{aligned}
x_i &= z_i \tan(\alpha_i), \\
y_i &= z_i \tan(\beta_i),
\end{aligned}
\end{equation}

where the horizontal and vertical viewing angles are given by

\begin{equation}
\begin{aligned}
\alpha_i &=
\left(c_i-\frac{N_c-1}{2}\right)
\frac{\Phi_x}{N_c-1}, \\
\beta_i &=
\left(r_i-\frac{N_r-1}{2}\right)
\frac{\Phi_y}{N_r-1}.
\end{aligned}
\end{equation}

Here, $r_i$ and $c_i$ are the row and column indices,
$N_r=10$, $N_c=16$, and $\Phi_x$, $\Phi_y$ are the calibrated fields
of view. The same $41^\circ\times52^\circ$ calibration is used for all
final 3-D reconstructions and plane fitting.

\subsection{Transient Feature Extraction and Reliability Estimation}

Each SPAD measurement provides a temporal photon histogram in addition
to the estimated range. Rather than treating all depth samples equally,
the proposed method derives a reliability value from three complementary
measurement cues: sensor confidence, transient peak amplitude, and peak
sharpness.

After background subtraction, peak amplitude is log-transformed and
robustly normalized, while sharpness is defined as one minus the
normalized FWHM of the dominant transient peak. The normalized feature
vector for point $i$ is

\begin{equation}
\mathbf{f}_i =
[C_i,\;P_i,\;S_i]^T,
\end{equation}

where $C_i$, $P_i$, and $S_i$ represent confidence, peak amplitude, and
peak sharpness, respectively. Measurement reliability is defined as

\begin{equation}
R_i =
w_C C_i + w_P P_i + w_S S_i,
\end{equation}

subject to

\begin{equation}
w_C+w_P+w_S=1,\qquad w_C,w_P,w_S \geq 0.
\end{equation}

\begin{figure}[t]
    \centering
    \includegraphics[width=\columnwidth]{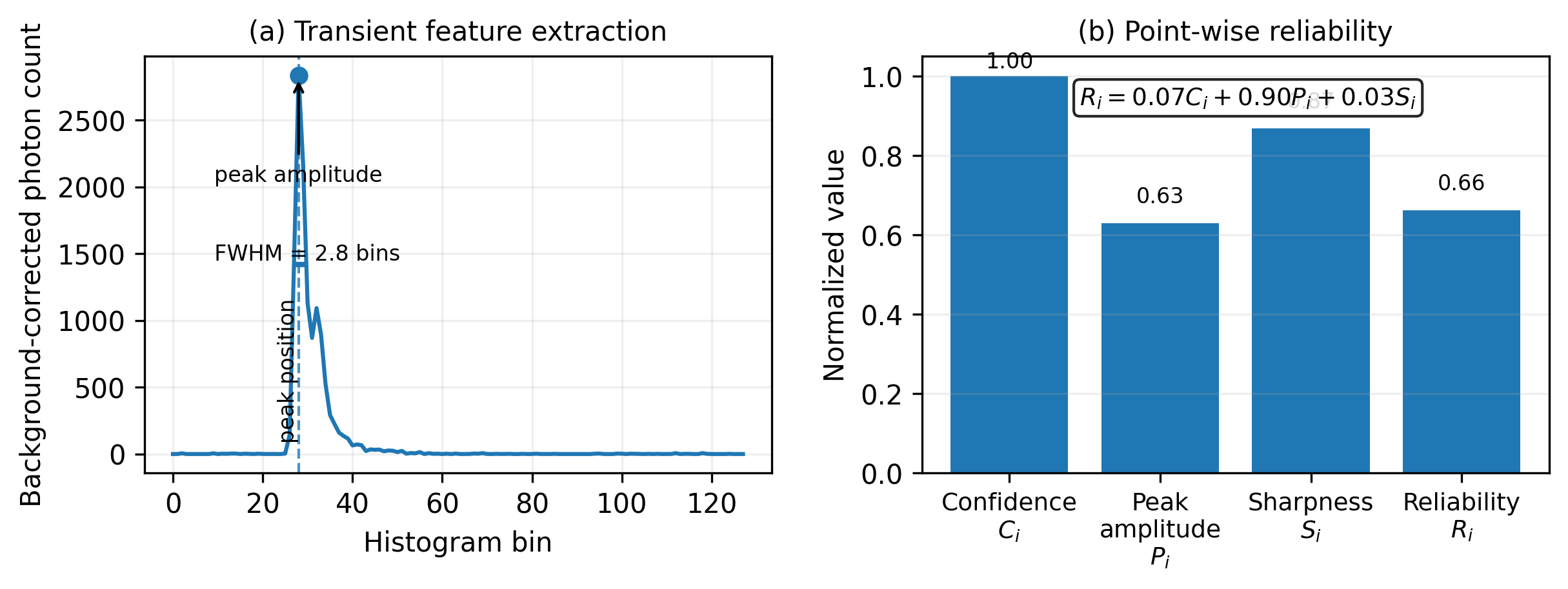}
    \caption{Transient feature extraction and point-wise reliability:
(a) representative transient histogram and (b) normalized features and
learned reliability $R_i$.}
    \label{fig:transient_reliability}
\end{figure}

The weights are optimized jointly with the reconstructed depth rather than
manually fixed. Fig.~\ref{fig:transient_reliability} illustrates the
transient features and resulting point-wise reliability.

For the representative $90^\circ$ experiment, the learned coefficients
are $(w_C,w_P,w_S)=(0.069,0.903,0.028)$, indicating a dominant
peak-amplitude contribution in this scene.

\subsection{Reliability-Aware Depth Optimization}

Let $z_i^{\mathrm{obs}}$ and $z_i$ denote the observed and optimized
depths of point $i$, respectively. The reconstructed depth is obtained
by minimizing

\begin{equation}
\mathcal{L} =
\lambda_d\mathcal{L}_{\mathrm{data}}
+\lambda_s\mathcal{L}_{\mathrm{smooth}}
+\lambda_p\mathcal{L}_{\mathrm{peak}}
+\lambda_{wp}\mathcal{L}_{wp}
+\lambda_{rm}\mathcal{L}_{rm}.
\end{equation}

The reliability-weighted data term is

\begin{equation}
\mathcal{L}_{\mathrm{data}} =
\frac{1}{N}
\sum_i
R_i\,
\rho
\left(
z_i-z_i^{\mathrm{obs}}
\right),
\end{equation}

where $\rho(t)=\sqrt{t^2+\epsilon^2}$ is the Charbonnier penalty; thus
lower-reliability samples can move farther from their observations.

Spatial consistency is imposed using a $k$-nearest-neighbor graph,

\begin{equation}
\mathcal{L}_{\mathrm{smooth}} =
\frac{1}{|\mathcal{E}|}
\sum_{(i,j)\in\mathcal{E}}
g_{ij}
\rho(z_i-z_j),
\end{equation}

where $g_{ij}$ is the transient-aware graph weight. $\mathcal{L}_{\mathrm{peak}}$ matches normalized optimized depth to
normalized peak position, $\mathcal{L}_{wp}$ penalizes deviation of the
learned reliability coefficients from $(0.4,0.4,0.2)$, and
$\mathcal{L}_{rm}$ preserves their initial mean reliability.

\subsection{V-Shaped Surface Geometry Estimation}

For quantitative evaluation, the reconstructed point cloud is separated
into the two dominant planar surfaces of the V-shaped target. Candidate
planes are estimated using robust fitting, with local point density and
spatial connectivity used to suppress sparse outliers.

The initial plane hypotheses follow a RANSAC-style robust model-fitting
strategy~\cite{fischler1981ransac}, after which density and connectivity
criteria favor spatially coherent support across each V-shaped face.

Each reconstructed plane is represented as

\begin{equation}
a x+b y+c z+d=0,
\end{equation}

with normal vector $\mathbf{n}=[a,b,c]^T$. The intersection direction
of the two planes is

\begin{equation}
\mathbf{l}
=
\mathbf{n}_1 \times \mathbf{n}_2.
\end{equation}

\begin{figure}[t]
    \centering
    \includegraphics[width=\columnwidth]{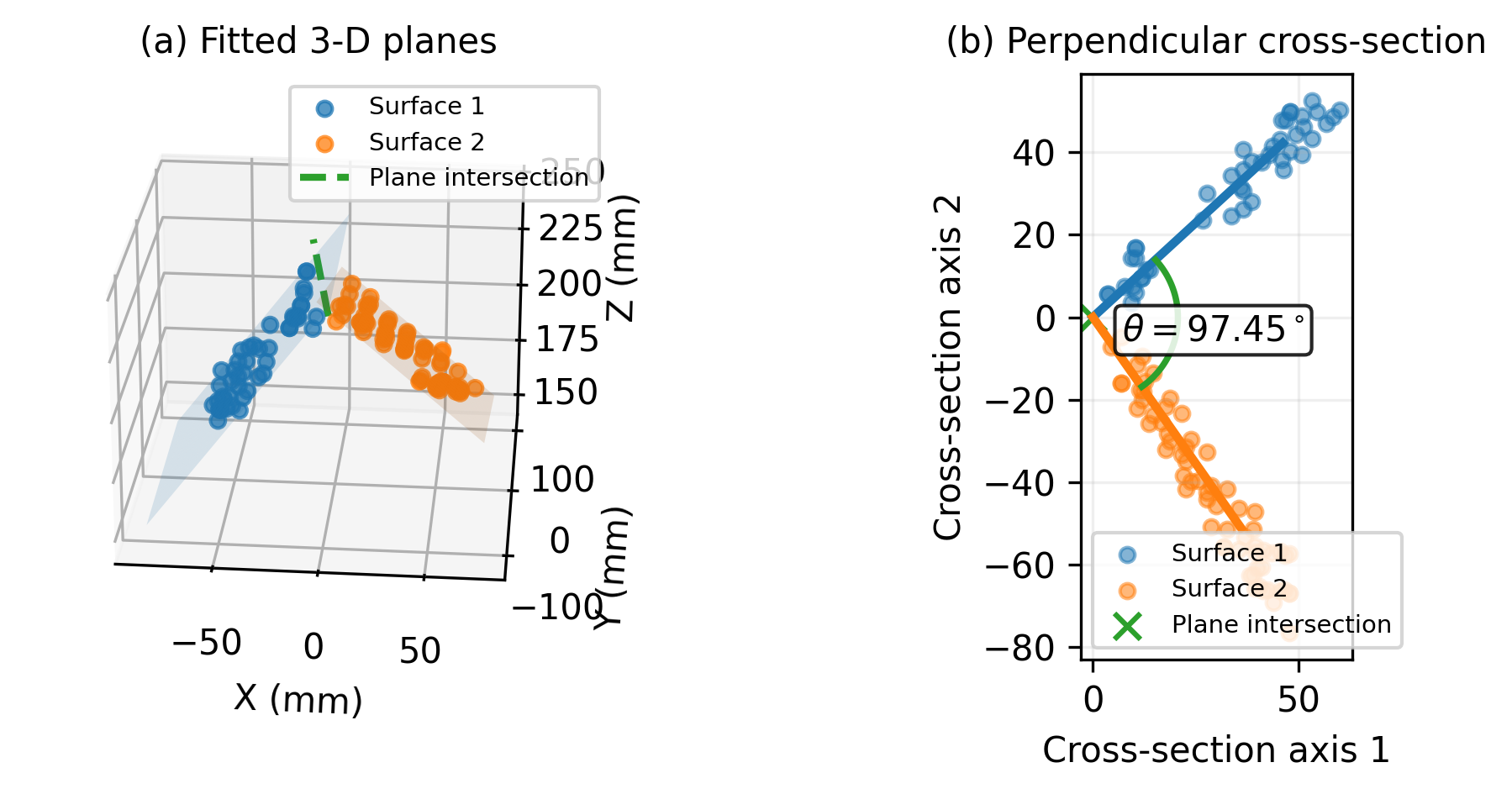}
    \caption{V-shaped geometry estimation: (a) fitted 3-D planes and
(b) perpendicular cross-section for the opening angle $\theta$.}
    \label{fig:v_geometry}
\end{figure}

A cross-sectional plane perpendicular to $\mathbf{l}$ is then used to
obtain the two surface directions. The V opening angle is computed from

\begin{equation}
\theta =
\cos^{-1}
\left(
\frac{\mathbf{d}_1^T\mathbf{d}_2}
{\|\mathbf{d}_1\|\|\mathbf{d}_2\|}
\right),
\end{equation}

where $\mathbf{d}_1$ and $\mathbf{d}_2$ denote the two oriented surface
directions in the cross-section. The geometry is illustrated in
Fig.~\ref{fig:v_geometry}.

\section{Experiments and Results}

\subsection{Experimental Setup}

Experiments were conducted using a TMF8821 multi-zone SPAD ToF sensor.
Three V-shaped targets with ground-truth opening angles of
$60^\circ$, $90^\circ$, and $120^\circ$ were evaluated. For each
target, transient measurements were acquired using two sensor
configurations and processed using identical reconstruction parameters.

\begin{figure}[t]
    \centering
    \includegraphics[width=0.8\columnwidth]{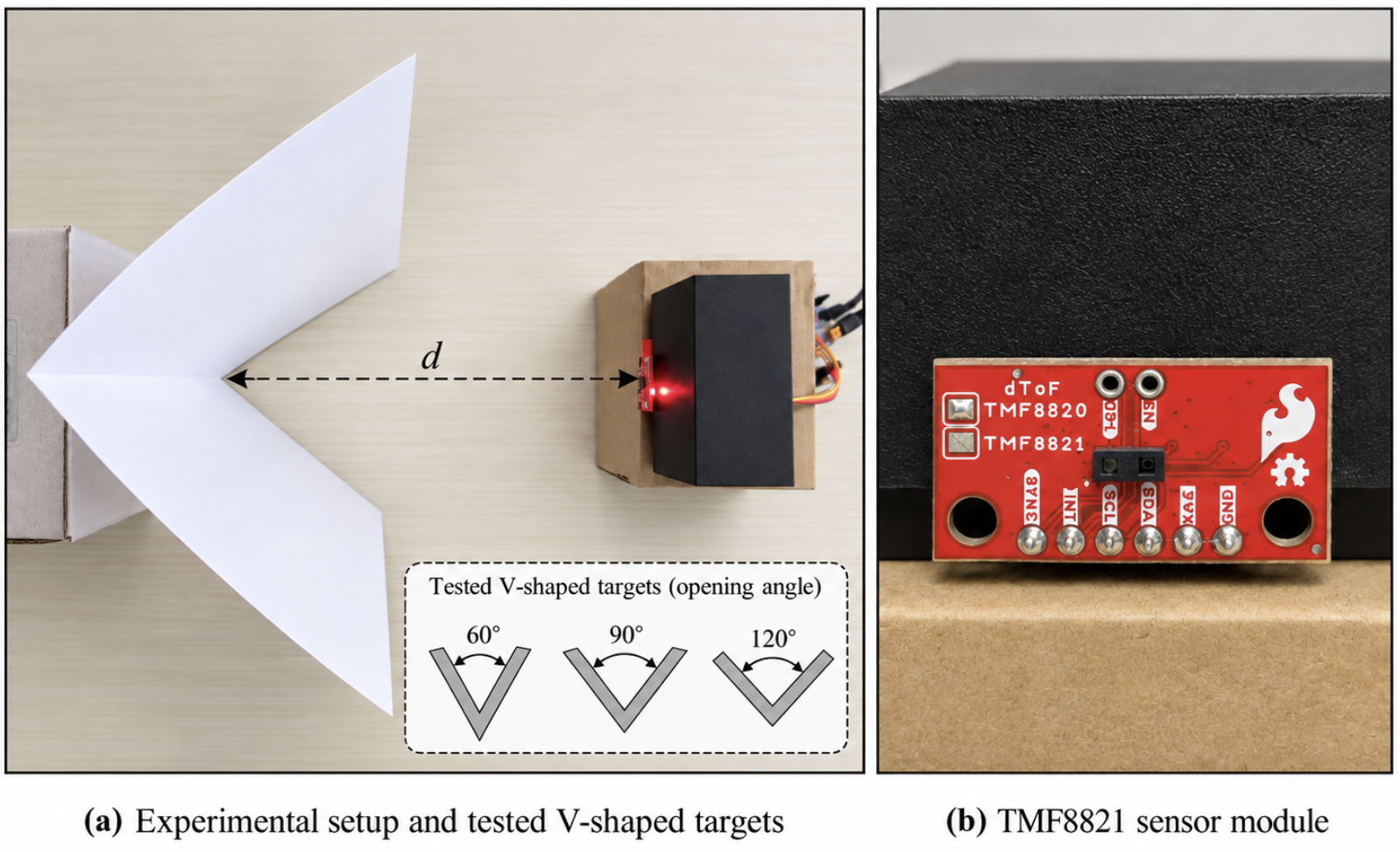}
    \caption{Experimental setup: (a) TMF8821-to-target arrangement and
(b) TMF8821 multi-zone dToF sensor module.}
    \label{fig:setup}
\end{figure}

The sensor field of view was calibrated to
$41^\circ\times52^\circ$ and kept fixed for all experiments.
No ground-truth angle information was used during reconstruction or plane
estimation. The setup and sensor module are shown in Fig.~\ref{fig:setup}.

All targets use 500 iterations with $(\lambda_d,\lambda_s,\lambda_p,\lambda_{wp},\lambda_{rm})=(1.0,0.8,0.03,0.02,0.10)$ and depth/weight
learning rates of 0.05/0.002; no target-specific tuning is used.

\subsection{V-Angle Reconstruction Accuracy}

The opening angle was estimated from the fitted 3-D planes using the
perpendicular cross-section of Section~II-D. Table~\ref{tab:angle}
summarizes the estimates and absolute errors before and after optimization.

\begin{table}[t]
\centering
\caption{V-angle reconstruction accuracy before and after optimization.}
\label{tab:angle}
\setlength{\tabcolsep}{3.5pt}
\begin{tabular}{c|cc|cc}
\hline
\multirow{2}{*}{GT} &
\multicolumn{2}{c|}{Before optimization} &
\multicolumn{2}{c}{After optimization} \\
 & Est. & Error & Est. & Error \\
\hline
$60^\circ$  & $66.42^\circ$  & $6.42^\circ$  & $65.21^\circ$  & $5.21^\circ$ \\
$90^\circ$  & $98.91^\circ$  & $8.91^\circ$  & $97.45^\circ$  & $7.45^\circ$ \\
$120^\circ$ & $138.29^\circ$ & $18.29^\circ$ & $136.31^\circ$ & $16.31^\circ$ \\
\hline
Mean & -- & $\mathbf{11.21^\circ}$ & -- & $\mathbf{9.66^\circ}$ \\
\hline
\end{tabular}
\end{table}

\begin{figure}[t]
    \centering
    \includegraphics[width=0.9\columnwidth]{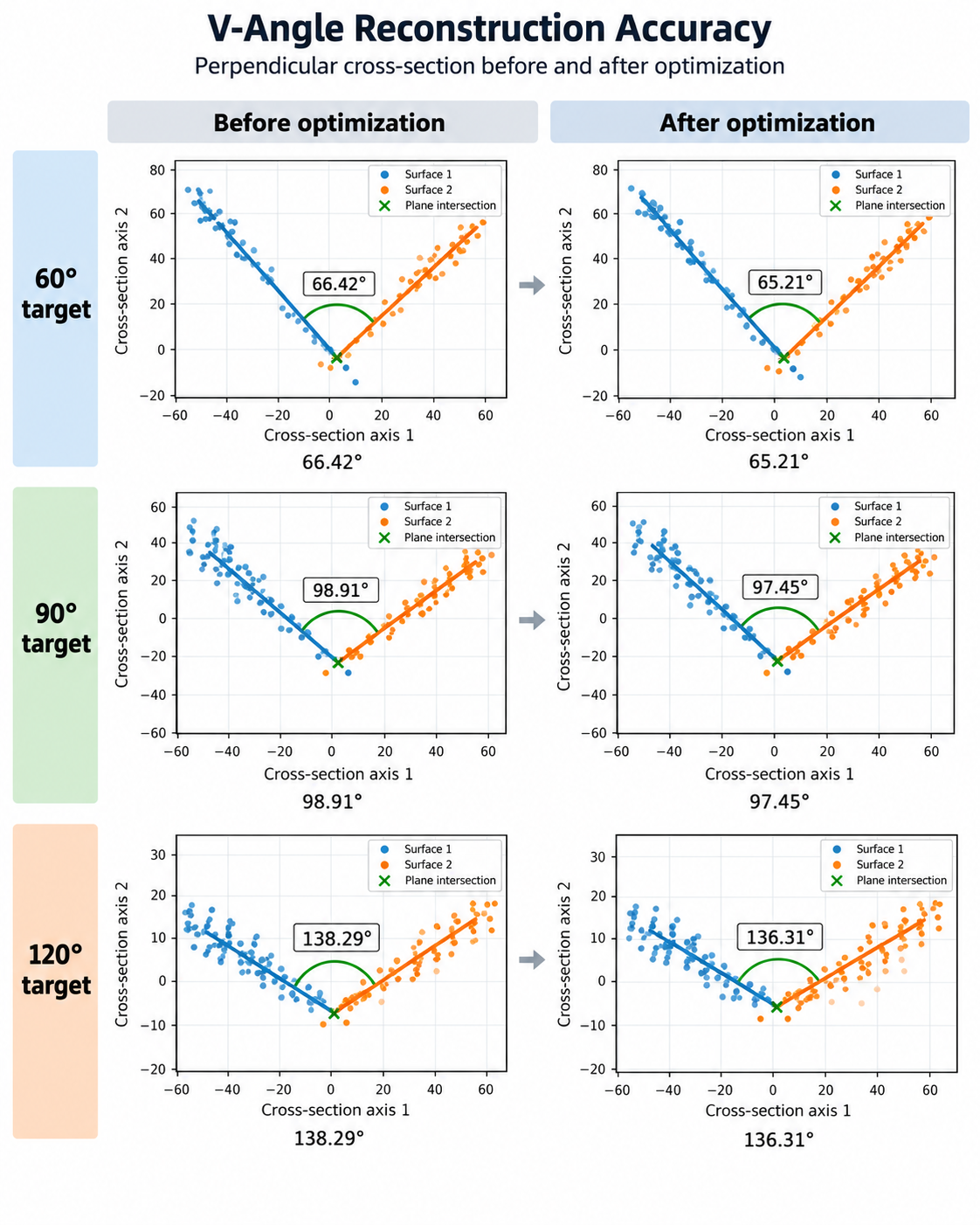}
    \caption{V-angle reconstruction before and after optimization for
    $60^\circ$, $90^\circ$, and $120^\circ$ targets.}
    \label{fig:angle_accuracy}
\end{figure}

The optimization reduces the angular error for all three targets, with
the mean error decreasing from $11.21^\circ$ to $9.66^\circ$ ($13.8\%$).
The $120^\circ$ target remains the most challenging case under sparse
spatial sampling. Fig.~\ref{fig:angle_accuracy} shows the corresponding
perpendicular cross-sections before and after optimization.

\subsection{Geometric Reconstruction Quality}

\begin{figure}[t]
    \centering
    \includegraphics[width=0.9\columnwidth]{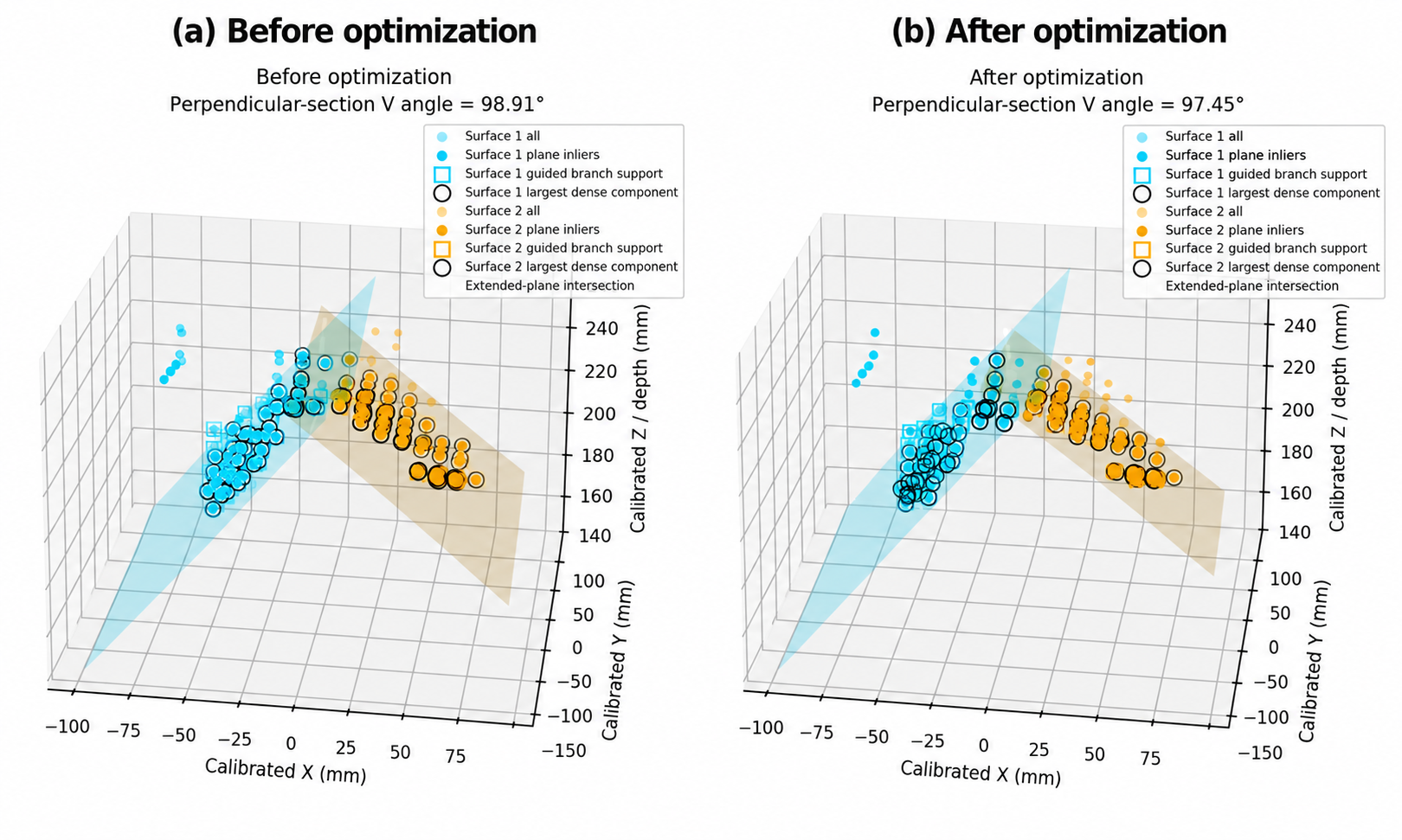}
    \caption{Geometric reconstruction quality for the $90^\circ$ V-shaped
    target before and after reliability-aware optimization.}
    \label{fig:geometry_quality}
\end{figure}

The $90^\circ$ target is used as a representative example of local planar
conformity. Fig.~\ref{fig:geometry_quality} compares the calibrated 3-D
reconstruction and fitted planes before and after optimization.

Geometric error is quantified by the point-to-plane RMSE,

\begin{equation}
\mathrm{RMSE}_{\mathrm{plane}}
=
\sqrt{\frac{1}{N}\sum_{i=1}^{N} d_i^2},
\end{equation}

where $d_i$ is the perpendicular point-to-plane distance; multi-peak
boundary measurements are excluded. The combined RMSE decreases from
$14.56$~mm to $13.65$~mm ($6.24\%$). Surface~1 decreases from
$19.06$~mm to $18.76$~mm, while Surface~2 decreases from $8.95$~mm to
$6.48$~mm, indicating improved local planar conformity.

\subsection{Effect of Multi-Configuration Acquisition}

\begin{figure}[t]
    \centering
    \includegraphics[width=0.9\columnwidth]{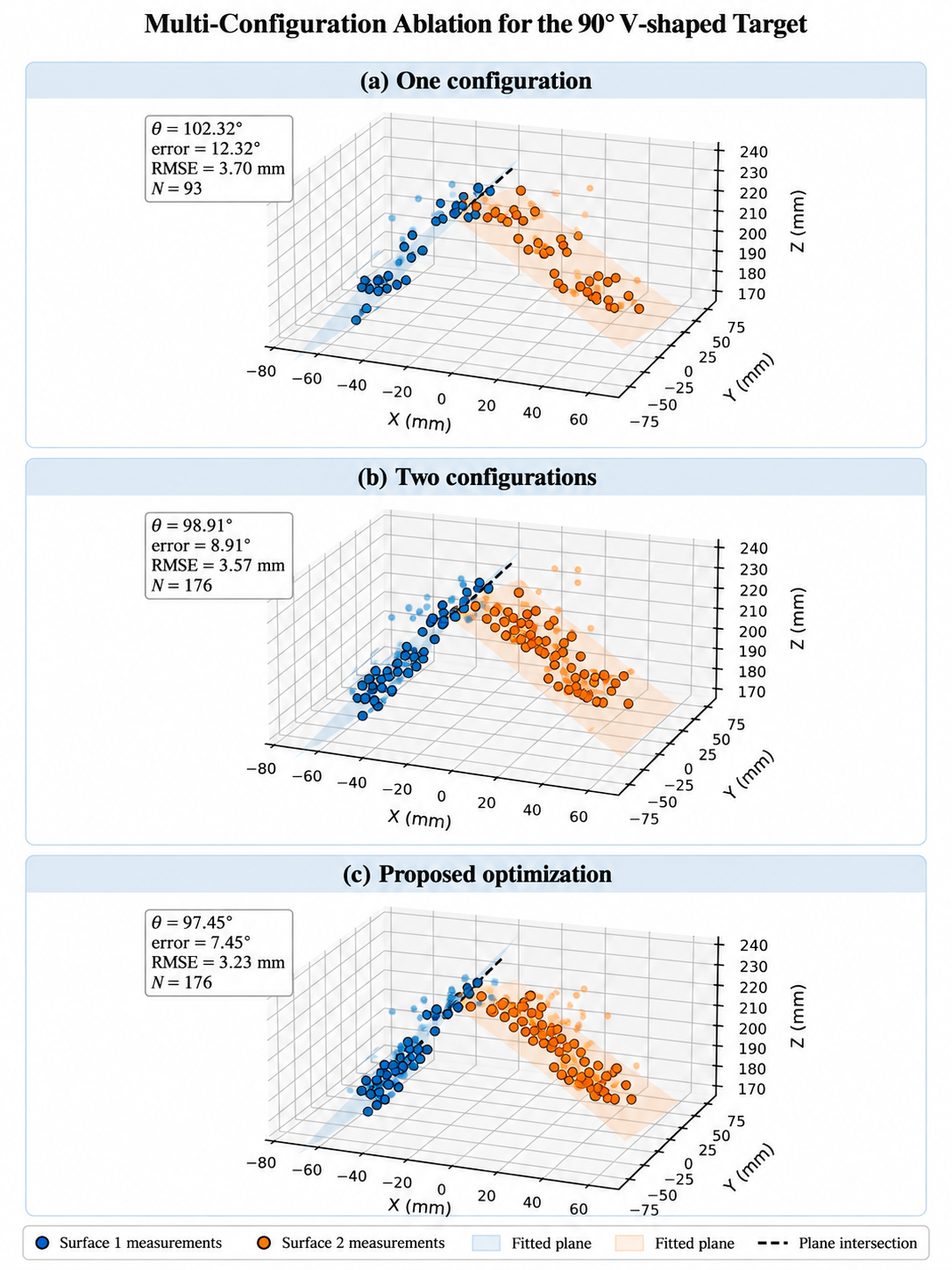}
    \caption{Ablation on the $90^\circ$ target: (a) one configuration
    before optimization, (b) two aggregated configurations before
    optimization, and (c) two configurations after the proposed
    reliability-aware optimization.}
    \label{fig:ablation}
\end{figure}

To isolate the contribution of complementary spatial sampling, the
$90^\circ$ target is additionally evaluated using Configuration B alone,
the aggregated B+C measurements before optimization, and the aggregated
B+C measurements after the proposed optimization. As shown in
Fig.~\ref{fig:ablation}, this comparison separates the effect of increased
spatial support from the subsequent reliability-aware depth refinement.

Table~\ref{tab:ablation} summarizes the same $90^\circ$ ablation
numerically. Using the second acquisition configuration increases the
retained spatial support from 93 to 176 points and reduces the angular
error from $12.32^\circ$ to $8.91^\circ$. The proposed reliability-aware
optimization further reduces the error to $7.45^\circ$.

\begin{table}[t]
\centering
\caption{Multi-configuration ablation on the $90^\circ$ target.}
\label{tab:ablation}
\setlength{\tabcolsep}{3.0pt}
\begin{tabular}{lccc}
\hline
Method & Points & Est. angle & Error \\
\hline
Config. B only & 93  & $102.32^\circ$ & $12.32^\circ$ \\
B+C raw        & 176 & $98.91^\circ$  & $8.91^\circ$ \\
B+C proposed   & 176 & $97.45^\circ$  & $7.45^\circ$ \\
\hline
\end{tabular}
\end{table}

\section{Conclusion}

A reliability-aware framework for V-shaped surface reconstruction using a
low-cost multi-zone SPAD sensor was presented. Across $60^\circ$,
$90^\circ$, and $120^\circ$ targets, the mean V-angle error decreased from
$11.21^\circ$ to $9.66^\circ$ ($13.8\%$); for the representative
$90^\circ$ target, point-to-plane RMSE decreased from $14.56$~mm to
$13.65$~mm. The results demonstrate the benefit of transient-derived
reliability for sparse SPAD reconstruction.

%
%

%
%



\bibliographystyle{IEEEtran}
\bibliography{references}

\end{document}